\documentclass{llncs}

\usepackage{url}
\usepackage{algorithm}
\usepackage{algpseudocode}
\usepackage{amsmath}
\usepackage{graphicx}
\usepackage[FIGTOPCAP]{subfigure}

\usepackage{fancyhdr}

\DeclareMathOperator{\argmax}{arg\,max}
\DeclareMathOperator{\match}{Match}
\DeclareMathOperator{\smatch}{SemMatch}
\DeclareMathOperator{\f}{f}

\begin{document}

\title{\vspace{3cm}\hspace{2cm}\\CBR with Commonsense Reasoning and Structure Mapping: An Application to Mediation}
\author{At{\i }l{\i }m G\"{u}ne\c{s} Baydin\inst{1,2} \and Ramon L\'{o}pez de M\'{a}ntaras\inst{1} \and Simeon Simoff\inst{3} \and Carles Sierra\inst{1}}
\institute{Artificial Intelligence Research Institute, IIIA\,-\,CSIC\\Campus Universitat Aut\`{o}noma de Barcelona, 08193 Bellaterra, Spain
\and
Escola d'Enginyeria\\Universitat Aut\`{o}noma de Barcelona, 08193 Bellaterra, Spain
\and
School of Computing and Mathematics\\University of Western Sydney, NSW 1797, Australia\\
\email{gunesbaydin@iiia.csic.es, mantaras@iiia.csic.es,\\s.simoff@uws.edu.au, sierra@iiia.csic.es}}

\maketitle 
\thispagestyle{fancy}
\lhead{This is an author-created pre-published version. The final publication is available at \url{www.springerlink.com}.\\Proc. 19th International Conference on Case-Based Reasoning, ICCBR 2011, Greenwich, London, September 12--15, 2011. LNCS (LNAI), vol. 6880, p. 378. Springer, Heidelberg (2011)}
\fancyheadoffset{2.5cm}
\fancyfoot{}

\begin{abstract}
Mediation is an important method in dispute resolution. We implement a case based reasoning approach to mediation integrating analogical and commonsense reasoning components that allow an \emph{artificial mediation agent} to satisfy requirements expected from a human mediator, in particular: utilizing experience with cases in different domains; and structurally transforming the set of issues for a better solution. We utilize a case structure based on ontologies reflecting the perceptions of the parties in dispute. The analogical reasoning component, employing the Structure Mapping Theory from psychology, provides a flexibility to respond innovatively in unusual circumstances, in contrast with conventional approaches confined into specialized problem domains. We aim to build a mediation case base incorporating real world instances ranging from interpersonal or intergroup disputes to international conflicts.
\end{abstract}

\section{Introduction} 
Mediation is a process of dispute resolution where an intermediary---called a mediator---assists two or more negotiating parties to reach an agreement in a conflict, who have failed to do so on their own. In the field of law, it is defined as a form of \emph{alternative dispute resolution} (ADR), i.e. a collection of techniques the parties might resort to instead of a judicial process, including, besides mediation, other types such as facilitation\footnote{The intermediary constructively organizes a discussion.} and arbitration\footnote{The intermediary has the power to impose a resolution.} \cite{Nabatchi2004}.

Two defining aspects of a mediation process are:
\begin{itemize}
\item that the mediators have special training that allows them to identify issues and explore options for solutions based on their experience, often by drawing parallels with similar past cases
\item and that the mediators handle the discussion with total impartiality, without having a personal stance on the discussed issues, and instead, offering to expand the discussion beyond the original dispute for allowing creative new solutions \cite{Pace2008}.
\end{itemize}

Within the field of artificial intelligence (AI), there is an active effort of research for studying negotiation processes using agent based modeling \cite{Kraus2001} and developing support tools for mediation \cite{Chalamish2007}. These, together with the recent formulation of a mediation framework by Simoff et al. \cite{Simoff2009}, provide a theoretical basis for a computational approach to mediation, which can promisingly address the aspects mentioned above.

In this paper, we describe the implementation of a case based reasoning (CBR) approach for an autonomous mediation system that can satisfy, to a sufficient degree, the requirements expected from a human mediator; and that can eventually tackle non-trivial disputes in a variety of problem domains. Our approach stems from, and is an improvement upon, the early case based problem solver nicknamed {MEDIATOR} by Simpson \cite{Simpson1985,Kolodner1989}. To this end, we introduce a CBR model that uses a case structure based on \emph{ontologies} and that incorporates: (1) a structure-matching \emph{analogical reasoning} component, which allows it to recall its experience with past cases in different domains; and (2) a \emph{commonsense reasoning} component, which emulates, to some extent, human-like innovation in reshaping the set of issues of conflict.

The role of analogical reasoning in the CBR algorithm that we present is twofold: it forms the basis of the retrieval stage with scores based on structural evaluation of possible analogies between case ontologies; and it is used in the adaptation stage for the inference of new knowledge about the current case by means of analogical mappings from retrieved cases. The commonsense reasoning component provides modifications of existing ontologies via commonsense knowledge, aiding in the uncovering of extensive analogies in all stages of the CBR algorithm.

After providing background information on analogical and commonsense reasoning in Section~\ref{SectionBackground}, we present the ideas underlying our approach by means of a structure mapping example in Section~\ref{SectionApproach}. Details of the implementation of our model are given in Section~\ref{SectionImplementation}, illustrated by a sample run of the CBR algorithm and the mediation process. This is followed by a discussion of building a mediation case base in Section~\ref{SectionCaseBase}. The paper ends with our conclusions and plans for future research in Section~\ref{SectionConclusion}.

\section{Background}
\label{SectionBackground}
CBR is a well-studied problem solving model in AI \cite{Kolodner1993,Aamodt1994,LopezDeMantaras2005,Rissland2006}, utilizing past experience in the form of a case base. By its nature, the CBR model can be viewed as a type of lazy, or instance-based, learning, i.e. without making any generalizations or deriving rules on how to solve the problems in the domain, in contrast with other models such as artificial neural networks and decision trees \cite{Leake1996,Chen2001}. Even if this lack of generalization can at times be seen as a disadvantage in particular application domains, the inherent instance-based approach of CBR renders it highly suitable as the backbone of a mediator. Due to the nature of the mediation process, instances of dispute can be conveniently represented as cases. Being kept intact for future reasoning by the CBR system, these cases also provide any decisions by the mediator with valuable explanation and backing as supporting precedents. 

The case representation structure that we use here is based on ontologies describing the perceptions of the negotiating parties on the set of issues forming the dispute; and we augment the conventional CBR cycle with analogical and commonsense reasoning modules operating on these ontologies, details of which are presented further below.

\subsection{Analogies and Structure Mapping}
Analogy is a cognitive process where information on an already known subject (the \emph{analogue} or \emph{base domain}) is transferred onto a newly encountered subject (the \emph{target domain}), through inference based on similarity. Analogical reasoning plays a crucial role in many defining aspects human cognitive capacity, such as problem solving, memory, and creativity \cite{Rumelhart1981,Gentner1983,Gentner1997}. There are several cognitive science approaches for the modeling of analogical reasoning \cite{French2002}, such as the \emph{Structure Mapping Theory} (SMT) \cite{Gentner1997} and a collection of other models inspired by it, for instance the work by Ferguson \cite{Ferguson1994} and Turney \cite{Turney2008}.

The very technique of CBR that we employ is occasionally emphasized as an analogy-based method, with an inherent facility of recalling instances from a case base that are similar to a given new situation. Nevertheless, in practice, CBR systems have been almost universally restricted to indexing and matching strategies as cases in a strictly defined single domain \cite{Aamodt1994}. This is arguably because of the difficulty in developing implementations capable of case retrieval by inter-domain analogies, and especially, adapting the solutions of past cases into the target domain. 

A pivotal part of our research is the integration of the computational implementation of SMT, the seminal \emph{Structure Mapping Engine} (SME) \cite{Falkenhainer1989,Gentner1997} into a CBR framework. By doing so, we achieve a degree of analogical reasoning ability for recalling analogous cases of past mediations in different base domains (CBR retrieval stage), together with the ability to bring new knowledge into the target domain by analogical inference (CBR adaptation stage). This is crucial for addressing the first requirement expected from a competent mediator mentioned before.

\subsection{Commonsense Reasoning}
Within AI, since the pioneering work by McCarthy \cite{McCarthy1958}, commonsense reasoning has been commonly regarded as a key ability that a system must possess in order to be considered truly intelligent \cite{Minsky2006}. There is an active effort to assemble and classify commonsense information involved in everyday human thinking into ontologies and present these to the use of scientific community in the form of commonsense knowledge bases, of which \emph{Cyc}\footnote{\url{http://www.cyc.com}} maintained by the Cycorp company and the \emph{ConceptNet} project\footnote{\url{http://csc.media.mit.edu}} of Massachusetts Institute of Technology (MIT) are the most prominent examples. The lexical database WordNet\footnote{\url{http://wordnet.princeton.edu}} maintained by the Cognitive Science Laboratory at Princeton University also has characteristics (via synonym/hypernym/hyponym relations) of a commonsense knowledge base.

The artificial mediator model implemented in this study interfaces with MIT ConceptNet and WordNet in the process of discovering middle-ground ontologies between the disputing parties and considering expansions or contractions of the involved ontologies to facilitate the analogical reasoning process, thereby allowing for solutions unforeseen before mediation.

\section{Approach: A Classical Example}
\label{SectionApproach}
\begin{figure}[t]
	\centering
	\includegraphics{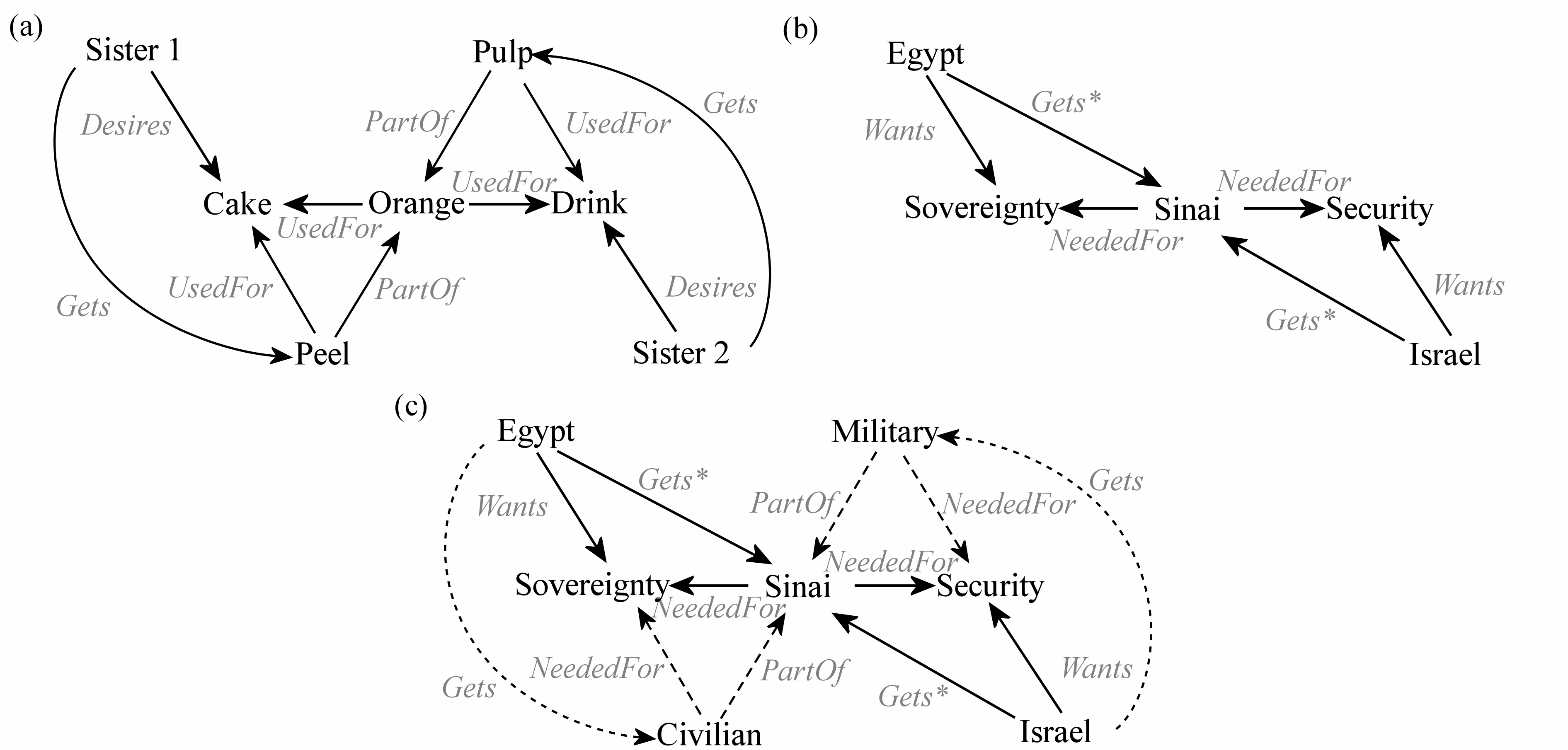}
	\caption{Ontologies of mediation cases in the orange dispute domain (a) and the Sinai peninsula dispute domain (b and c). Vertices and edges respectively represent concepts and relations; dashed edges represent subsequently discovered relations leading to a solution; relations causing conflict are marked with ``*''.}
	\label{FigureApproachExample}
\end{figure}

\emph{The orange dispute and Sinai Peninsula}: To illustrate our approach to mediation and analogical reasoning, let us briefly describe a classical mediation example that embodies the essence of our approach, which was introduced by Simpson \cite{Simpson1985} and later used by Kolodner \cite{Kolodner1989,Kolodner1993} in her seminal work on CBR. Considering a resource dispute where two sisters want the same {\tt orange} (Figure~\ref{FigureApproachExample}(a)), a mediator first assumes that a simple division of the {\tt orange} into two would solve the dispute, but this is unacceptable for the parties. After a point in the mediation process, it is revealed that one sister wants the {\tt orange} for the reason of cooking a {\tt cake} (for which its {\tt peel} is sufficient) and the other for making a {\tt drink} (for which its {\tt pulp} is sufficient). The solution is then to redefine the disputed resource as an entity composed of a {\tt pulp} and a {\tt peel} and to assign these to the parties (dashed edges).

This simple mediation case is strikingly similar to a real world crisis in international relations, where the countries of {\tt Egypt} and {\tt Israel} had a dispute over the control of the {\tt Sinai} peninsula following the Yom Kippur War in 1973 (Figure~\ref{FigureApproachExample}(b)). Starting the mediation with the initial ontology (lacking the dashed edges), through analogical reasoning by SME, we can consider a structural correspondence between concept pairs in these domains (e.g.  {\tt orange}--{\tt Sinai}). Moreover, by this analogical mapping we can \emph{infer} that, corresponding to {\tt pulp} and {\tt peel} in the base domain, there may exist two more concepts in the target domain that we can base a solution upon (linked by dashed edges in Figure~\ref{FigureApproachExample}(c)), which incidentally corresponds to a simplistic view of how the dispute was successfully mediated by the US president Jimmy Carter in 1979. The uncovering of these postulated concepts---namely, that the control of a territory has {\tt military} and {\tt civilian} aspects---is addressed by the commonsense reasoning module of our approach using ConceptNet. This works by generating expansions of the ontology at hand by a given factor---by attaching more concepts to existing concepts through identified commonsense relations---and making use of the robustness of SME to capture relations and concepts relevant to the considered analogy. For overcoming possibly different semantics used in naming the relations in the base and target domains (note the {\tt desires}--{\tt wants}, {\tt usedFor}--{\tt neededFor} relations in Figure~\ref{FigureApproachExample}(a) and (b)), we make use of WordNet by considering the synsets\footnote{A \emph{synset} or \emph{synonym ring} is a set of synonyms that are interchangeable without changing the truth value of any propositions in which they are embedded.} in the matching of relation structures by SME.

This solution by ``agreeable division based on the real goals of the disputants'' can form a basis for solving many future cases of mediation involving resource allocation. Note that, by using an analogical reasoning approach, this ability is maintained regardless of the semantics of problem domains, because we reach solutions only through similarities in the ontological structures of disputes.

\section{Implementation of the Mediator}
\label{SectionImplementation}
The fundamental part underlying our approach is a CBR cycle integrated with analogical and commonsense reasoning components, capable of: (1) creating a middle-ground ontology representing the views of all agents in dispute; (2) using this ontology for the retrieval of cases through analogical reasoning from a case base of previous successful mediations in various domains; and (3) adapting a solution for the current case, again by utilizing the middle-ground ontology and the ontology of the retrieved previous case, taking the goals and reservations of the parties into account.

\begin{figure}[t]
	\centering
	\includegraphics{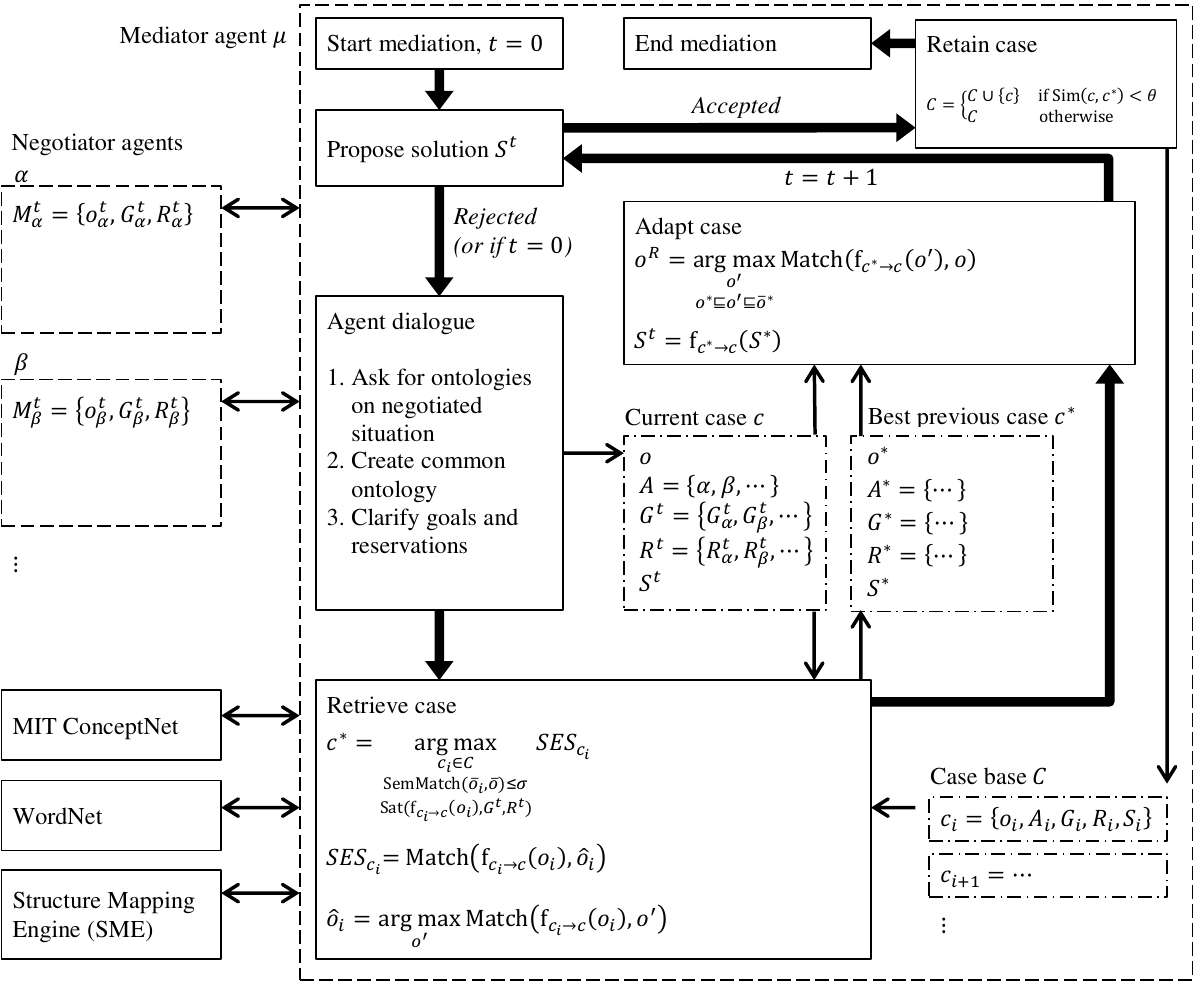}
	\caption{Architecture of the implemented CBR model and the flow of the mediation process. Thick arrows represent the flow of CBR algorithm while thinner arrows represent the utilization of case data. Bidirectional arrows represent communication between components.}
	\label{FigureMediatorModel}
\end{figure}

This system is to act as a mediator in-between several negotiating agents in a multiagent environment, in similar fashion to the ``curious negotiator'' model by Simoff et al. \cite{Simoff2002}. The following sections give a description of the model and its implementation together with the flow of the mediation process in Figure~\ref{FigureMediatorModel} and an example run in Table~\ref{TableMediationExample}.

\begin{table}[!th]
	\centering
	\caption{Example mediation process with the implementation. Relations between concepts are presented in LISP notation; relations causing conflict are marked with ``*''.}
	\label{TableMediationExample}
	\includegraphics{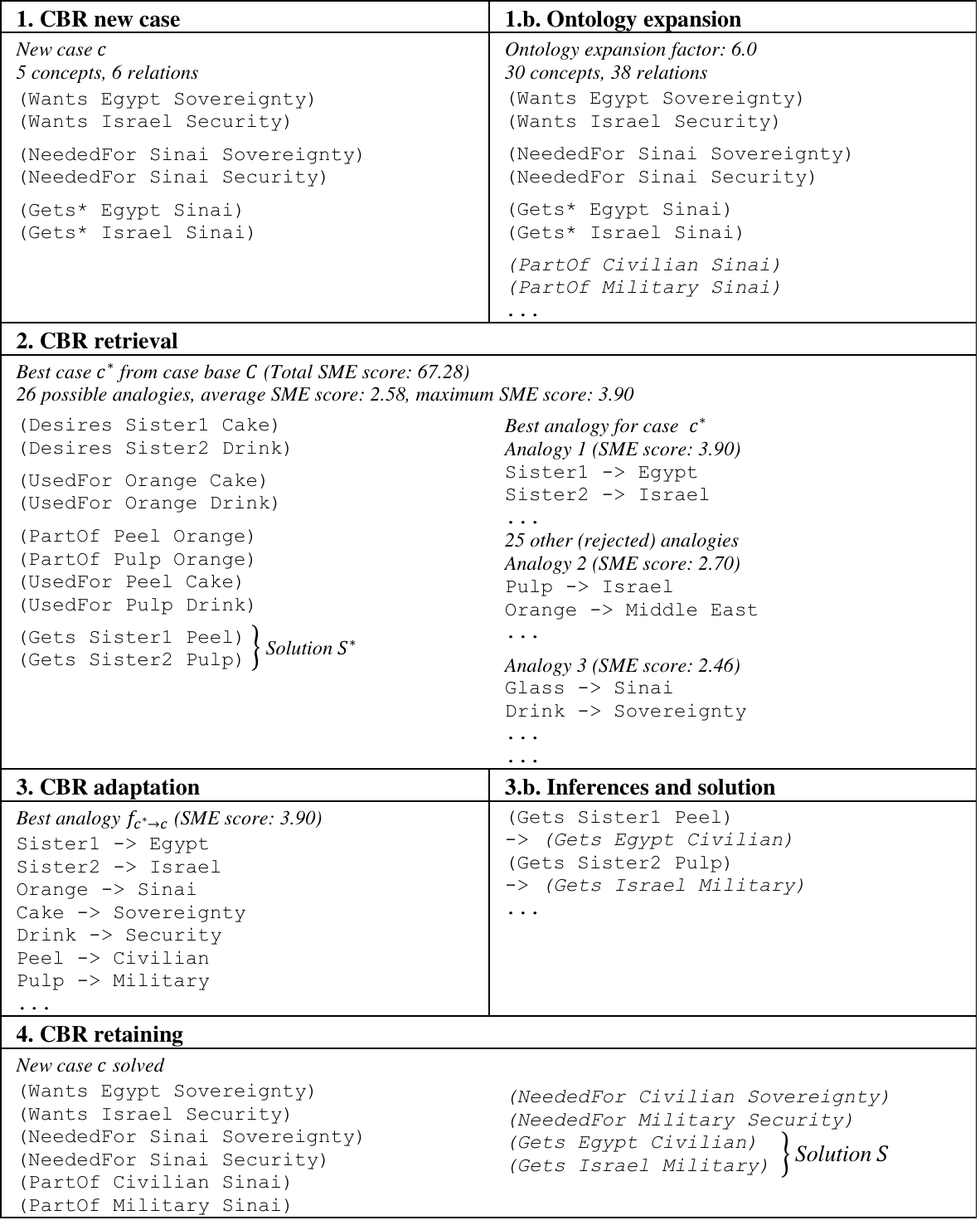}
\end{table}

\subsection{Case Representation}
After initial dialogue between the mediator and the agents in dispute, a newly acquired dispute $c$ is represented in the same manner as the cases in the CBR case base, except that the dispute at hand will lack a solution. The case base $C$ holds instances of successfully ended past mediations. Each case $c_i$ in the case base is fully described by the set 

\begin{equation}
{c_i}=\{o_i, A_i, G_i, R_i, S_i\}\enspace{,}
\end{equation}

\noindent denoting respectively the associated ontology of the dispute, the agents, their goals, their reservations, and the solution (Figure~\ref{FigureMediatorModel}). Even if $A_i$, $G_i$, $R_i$, and $S_i$ already exist as subgraphs of concepts and relations embedded into the ontology $o_i$ (Figure~\ref{FigureApproachExample}), they are also explicitly listed as features of the case $c_i$ for indicating which concepts and relations correspond to the agents together with their goals and reservations, and also for case indexing purposes. We permit the possibility that the parties modify their stances (e.g. $M_\alpha^t=\{o_\alpha^t,G_\alpha^t,R_\alpha^t\}$ in Figure~\ref{FigureMediatorModel}) after successive solution proposals $S^t$, where $t$ is the time index of the number of CBR cycles in the current run.

\subsubsection{Commonsense Reasoning Module}
\begin{figure}[t]
	\centering
	\subfigure[\hspace{10cm}]{
		\includegraphics[width=7cm]{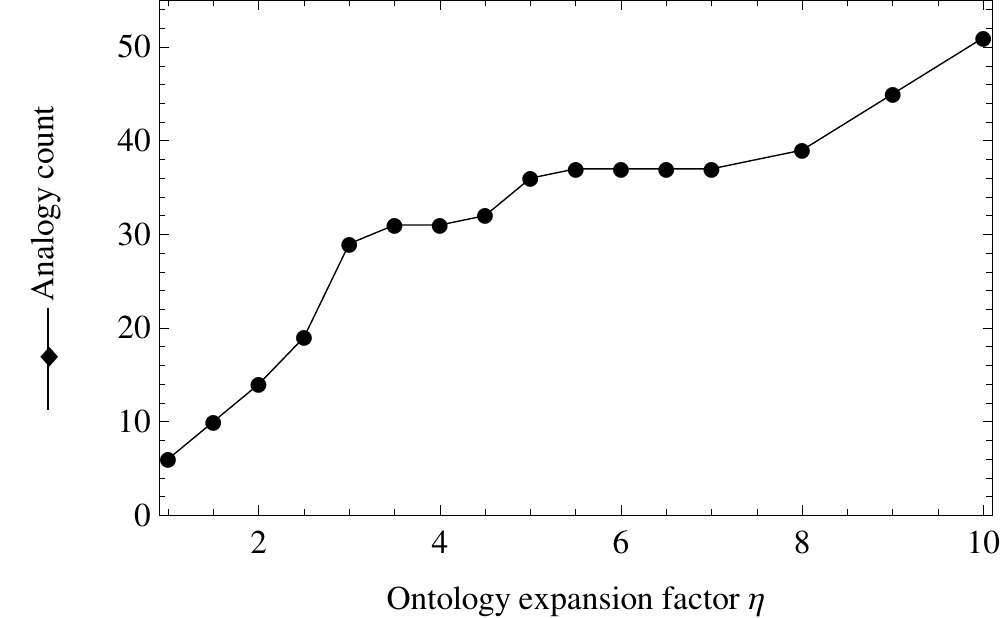}
		\label{FigureExpansionA}
	}
	\subfigure[\hspace{10cm}]{
		\includegraphics[width=7cm]{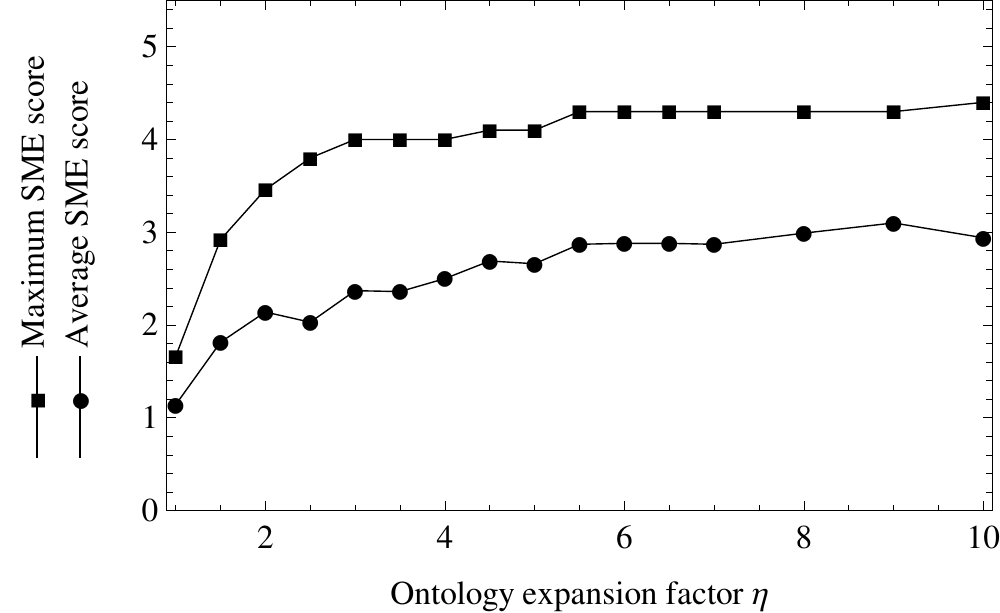}
		\label{FigureExpansionB}
	}
	\caption{Plot of the number of analogies (a) and the maximum and average SME structural evaluation scores (b) corresponding to a given ontology expansion factor $\eta$, computed for the orange dispute domain.}
	\label{FigureExpansion}
\end{figure}

For enabling the discovery of extensive analogies between different domains, we treat every given ontology $o$ as a partial view of a more general ontology $\bar{o}$, denoted $o\,{\sqsubseteq}\,\bar{o}$. We produce expansions of a given ontology (e.g. $o\,{\sqsubseteq}\,{o^\prime}\,{\sqsubseteq}\,\bar{o}$ in Figure~\ref{FigureMediatorModel}) by inserting into it new concepts and relations involving existing concepts, until the total number of concepts equals its previous value multiplied by a given expansion factor $\eta\,{\geq}\,1$ (Algorithm~\ref{AlgorithmOntologyExpansion}). Figure~\ref{FigureExpansion} gives an example, for the case of orange--Sinai peninsula analogy, of how the number of discovered analogies and the average and maximum structural evaluation scores are affected by the ontology expansion factor. As illustrated by this example, it is generally observed that there exists an asymptotic upper bound for the quality of attainable analogies between two domains (Figure~\ref{FigureExpansion}(b)). We therefore reason that, while the number of analogies keeps monotonically increasing with the expansion factor $\eta$ (Figure~\ref{FigureExpansion}(a)), the best analogy does not improve further after its maximum possible extent is uncovered by the expansions up to that point. Based on this observation, we limit the expansion factor $\eta$ in our implementation by a maximum ${\eta_{max}}=6$.

\begin{algorithm}
\caption{Ontology expansion}
\label{AlgorithmOntologyExpansion}
\begin{algorithmic}
\Procedure{Expand}{$o,\eta$}\Comment{Ontology $o$, expansion factor $\eta$}
	\State Set $n={\lfloor}(\eta-1)\operatorname{NumConcepts}(o){\rfloor}$\Comment{Number of new concepts}
	\While{$n>0$}
		\State Select $con=Random(o)$\Comment{Random concept in $o$}
		\State Create ontology $r$ of all concepts in relation with $con$
		\State \;\;from commonsense knowledge bases
		\If{$r\neq\emptyset$}
			\State Select $new=Random(r)$\Comment{Random concept in $r$}
			\State $\operatorname{Append}(o,new)$\Comment{Append $new$ to $o$ with corresponding relation}
			\State $n=n-1$\Comment{New concept appended}
		\EndIf
	\EndWhile
\EndProcedure
\end{algorithmic}
\end{algorithm}

As the commonsense knowledge base, our implementation depends mainly on ConceptNet \cite{Havasi2007}, part of the ``Common Sense Computing Initiative'' framework of the MIT Media Lab.\footnote{\url{http://csc.media.mit.edu/}} This commonsense knowledge base is being built by the contributions of a few thousand people across the world and is maintained as a simple graph of concepts and binary relations. We also make use of WordNet \cite{Fellbaum1998} as a commonsense knowledge base, given that, in addition to grouping words into synsets (e.g. ``object, thing, article, item, entity''), it also provides a taxonomical structure defined by hyponym and hypernym relations (e.g. ``dog''--``canine''--``mammal''--``vertebrate''--``animal'') that are highly useful and noise-free as compared to ConceptNet.

The implementation accesses MIT ConceptNet 4.0 as an XML service through a REST application interface\footnote{\url{http://csc.media.mit.edu/docs/conceptnet/webapi.html}} and WordNet 2.1 from local database files. In further stages of research, we plan to add to these the proprietary knowledge base Cyc \cite{Lenat1995}, a portion of which has been recently released as an open source project called OpenCyc;\footnote{\url{http://www.cyc.com/cyc}} and information mining agents, which would crawl online textual information for specific pieces of ad hoc knowledge and deliver these to the mediation agent in a structured manner. This could be of help especially in the adaptation stage of CBR.

\subsection{Retrieval}
The majority of CBR systems have case specifications consisting of preselected attributes and values; and use techniques such as nearest neighbor computations or decision trees for retrieval \cite{Cunningham1998}. In contrast to this, here, case retrieval is a complex task based on the structural and semantic composition of ontologies associated with cases, using SME to find cases analogous to the current dispute and not necessarily in the same problem domain. 

During the retrieval process, presented in Algorithm~\ref{AlgorithmRetrieval}, the best case $c^*$ is selected as the case maximizing the structural evaluation score 

\begin{equation}
SES_{c_i}=Match(f_{{c_i}{\rightarrow}c}(o_i),\hat{o}_i)\enspace{,}
\end{equation}

\noindent between the current case $c$ and cases $c_i$, where $f_{{c_i}{\rightarrow}c}(o_i)$ is the analogical mapping of $o_i$ from the domain of $c_i$ to that of $c$, and $\hat{o}_i$ is an expansion of ontology $o$ for comparison with $o_i$. The expansion $\hat{o}_i$, in turn, is found by expanding $o$ by the commonsense reasoning module 

\begin{equation}
\hat{o}_i=\underset{\underset{o{\sqsubseteq}o^\prime{\sqsubseteq}\bar{o}}{o^\prime}}{\argmax} \match(\f_{{c_i}{\rightarrow}c}(o_i),o^\prime)\enspace{,}
\end{equation}

\noindent as to maximize its match with $f_{{c_i}{\rightarrow}c}(o_i)$ (Algorithm~\ref{AlgorithmOntologyExpansion}). The two constraints on the considered cases $c_i$, 

\begin{equation}
\begin{split}
 \operatorname{Sat}(\f_{{c_i}{\rightarrow}c}(o_i),{G^t},{R^t})\\
 \smatch({\bar{o}_i},\bar{o})\leq\sigma\enspace{,}
 \end{split}
 \end{equation}
 
 \noindent ensure that the selected case permits a mapping satisfying the goals $G^t$ and reservations $R^t$ in the current case and that the general ontologies $\bar{o}_i$ and $\bar{o}$ lie in sufficiently different domains (i.e. the concepts they include are semantically dissimilar compared with a treshold $\sigma$).

\begin{algorithm}
\caption{Case retrieval}
\label{AlgorithmRetrieval}
\begin{algorithmic}
\Procedure{Retrieve}{$c,C$}\Comment{Current case $c$, case base $C$}
	\For{each case $c_i$ in the case base $C$}
		\State Compute $\hat{o}_i=\underset{\underset{o{\sqsubseteq}o^\prime{\sqsubseteq}\bar{o}}{o^\prime}}{\argmax} \match(\f_{{c_i}{\rightarrow}c}(o_i),o^\prime)$\Comment{Expansions $\hat{o}_i$ of ontology $o$}
		\State Compute ${SES_c}_i=\match(\f_{{c_i}{\rightarrow}c}(o_i),\hat{o}_i)$\Comment{Structural evaluation scores}
	\EndFor
	\State Select $c^* = \underset{\underset{\operatorname{Sat}(\f_{{c_i}{\rightarrow}c}(o_i),{G^t},{R^t})}{\underset{\smatch({\bar{o}_i},\bar{o})\leq\sigma}{{c_i}{\in}C}}}{\argmax}\;{SES_c}_i$
	\State \textbf{return} $c^*$\Comment{Case with best matching}
\EndProcedure
\end{algorithmic}
\end{algorithm}

SME is very robust and quick with the computation of analogical matchings between ontologies in real time. For instance, the discovery of the 26 possible analogies between the query case and the retrieved best case given in Table~\ref{TableMediationExample} takes only a fraction of a second on a currently average laptop computer. This is achieved by---instead of computing every possible mapping between two ontology graphs---using an incremental procedure for combining local matches into global match hypotheses under heuristic rules warranting structural consistency \cite{Falkenhainer1989}. Still, in the event that the mediation case base becomes prohibitively large for the computation of structural evaluation scores for each retrieval phase, a \emph{base filtering} approach for retrieval \cite{Smyth1993} can also be employed, in effect running the analogical reasoning process on a smaller subgroup for each retrieval.

\subsubsection{Analogy Module}
For analogical reasoning between different domains, we employ our own implementation of SME as described by Falkenhainer et al.\cite{Falkenhainer1989}, a very fast analogical matching algorithm derived from SMT with a firm basis in cognitive science and often cited as the most influential work on computational analogy-making \cite{Turney2008}. In addition to computing the match score between two ontologies (e.g. $Match(f_{{c_i}{\rightarrow}c}(o_i),o^\prime)$), SME also provides the mapping function $f$ between the domains, through which one can infer previously unknown information in the target domain ontology (e.g. $f_{{c^*}{\rightarrow}c}(S^*)$) (Figure~\ref{FigureMediatorModel}).

Given two ontologies in different domains, SME gives a set of all structurally meaningful analogical mappings between these, each with its attached structural evaluation score. While we pick the analogy with the highest score as the basis for the analogical mapping function $f$, in our implementation of the $Match$ function, we sum up the scores from all possible analogies between the given two ontologies as a measure of the susceptibility of these two to analogies (Table~\ref{TableMediationExample}, retrieval).

The discovery of analogies between ontologies by SME is guided by the structure of relations between concepts and this is dependent upon a consistent labeling of the types of relations across these ontologies. As it is highly probable that different semantics for the naming of the same relations would be used when the ontologies belong to different domains (Figure~\ref{FigureApproachExample}), our model makes use of WordNet to attach a tag of the synset of each relation within the ontologies. The SME matching then operates between these instead of the particular name of each relation.

\subsection{Adaptation}
In principle, the adaptation stage of our implementation falls under \emph{substitutional adaptation}, where the substitutions are made by the analogical mapping function $f$ from $c^*$ to $c$. Hence, we get a candidate solution to the current case by the mapping 

\begin{equation}
f_{{c^*}{\rightarrow}c}(S^*)\enspace{,}
\end{equation}

\noindent where $S^*$ is the solution of the retrieved case $c^*$ (Table~\ref{TableMediationExample}, adaptation). We use the mapping $f_{{c^*}{\rightarrow}c}$ corresponding to an analogical match established between $o$, the ontology of the current case, and 

\begin{equation}
{o^R}=\underset{\underset{{o^*}{\sqsubseteq}o^\prime{\sqsubseteq}{\bar{o}^*}}{o^\prime}}{\argmax} \match(\f_{{c^*}{\rightarrow}c}(o^\prime),o)\enspace{,}
\end{equation}

\noindent an expanded ontology of the retrieved case (Figure~\ref{FigureMediatorModel}).

A possibility that we consider for a more powerful adaptation stage is to use a \emph{generative adaptation} technique, where each of the derivational steps of solution from the domain of the retrieved case would be mapped into the domain of the current case, and the solution would be reached by reusing and modifying these steps for the current case. For this, the case structure has to be modified to include the solution steps in addition to the solution arrived at.

\subsection{Retaining}
At the point in the CBR cycle where the proposed solution is accepted by the parties in dispute, the case base $C$ is updated to include the case $c$ now with an accepted solution $S^t$. This new solution is retained whenever the newly solved case $c$ is sufficiently different from the retrieved case $c^*$, compared with a similarity threshold parameter $\theta$, in order to prevent overpopulation of the case base with instances of essentially the same dispute (Figure~\ref{FigureMediatorModel}).

\section{Building a Mediation Case Base}
\label{SectionCaseBase}
Instances of conflict cited in mediation literature range from familial disputes about inheritance or divorce to workplace disputes between coworkers, and from tenant--landlord disputes about the rent of a property to full-fledged armed conflicts between countries \cite{Domenici2001}. Even if these pose an apparent diversity, we argue that there should be a limited number of supercategories of conflicts subject to mediation---where each given conflict will be analogous to all other conflicts within its category---such as resource allocation, compensation, or scheduling. In fact, our analogical reasoning approach can be extended to discover these categories in a supplied case base in an unsupervised manner, via clustering, using the SME structural match scores as a distance metric.
 
\subsection{International Conflict Databases}
An important topic within mediation studies is international conflict resolution. As already exemplified in Figure~\ref{FigureApproachExample}, it is reasonable that there exists enough structural similarity between seemingly unrelated domains such as familial disputes and the resolution of international conflicts, which would allow our approach to uncover meaningful analogies. Incorporating international conflicts into the case base is desirable for benefiting from experience with non-trivial real world disputes and also rendering our research interesting from the perspective of social scientists in international relations and related fields.

There are several efforts for cataloging international conflicts, such as the \emph{Confman} database \cite{Bercovitch1993}, the \emph{International Crisis Behavior} (ICB) project \cite{Brecher2000}, and the \emph{Uppsala Conflict Data Program} (UCDP) maintained by Uppsala University in Sweden and the International Peace Research Institute (PRIO) in Oslo, Norway. In particular, the UCDP Peace Agreement Dataset v. 1.0, 1989--2005 \cite{Harbom2006}, which provides information on third party involvement in peace negotiations, together with the Confman database of conflict management attempts during 1945--1989, provide the best resources for our purpose.

On the other hand, as the main aim of these datasets is to index and classify conflicts according to a chosen set of features, they do not wholly submit to our approach due to a lack of descriptions of the steps in the mediation process.

\subsection{Case Generation}
Recognizing (1) the near absence of mediation knowledge bases that include the exact steps of mediation in each case and that contain sufficiently detailed information enabling the formulation of ontologies, and (2) that mediation-prone disputes fall into an arguably limited number of categories, we consider an approach for generating mediation cases with metaheuristic optimization techniques. We propose for future work using \emph{genetic programming} (GP)\footnote{An evolutionary optimization method working on tree structures.} \cite{Koza2003} for creating large numbers of ontologies that are analogous to a given ontology, guided by a fitness function based on SME structural evaluation scores. The tree-based based nature of GP makes it highly suitable for this purpose, with appropriate modifications taking the unconnected nature of ontology graphs into account.

\section{Conclusion}
\label{SectionConclusion}
We have presented a case based artificial mediator implementation integrating analogical and commonsense reasoning. The components of the model work together to create a system with the ability to back any of its solutions with supporting explanations, in terms of analogies with prior cases in its case base. This feature is highly advantageous within the context of law, where reference to precedent cases are deemed highly important.

In terms of practical value, the line of research following from this study has potential to find real life applications in diverse domains involving negotiation, among which law, dispute resolution, international conflict management, and commerce are foremost. These can be in the form of a support system augmenting the abilities of a human mediator, as well as in some situations replacing the human component altogether. For the case of law, this study can form a meaningful connection with several existing research efforts in the field of AI and law, such as the using of analogies in legal problem solving \cite{Branting2000} and ethical reasoning \cite{McLaren1999}.

In future work we plan to address further development of the mediation case base, largely by the case generation technique we mentioned; and improving the adaptation stage of our model, by generative adaptation. Another issue that we will concentrate on is the process of dialogue between the agents and the mediator (represented by the three steps before retrieval in Figure~\ref{FigureMediatorModel}). Rather than taking the ontological perceptions of agents as given, this dialogue should be implemented from an AI argumentation perspective. Lastly, it would be certainly interesting to see our integration of structure mapping and commonsense reasoning in a CBR framework applied to problem domains other than mediation.

\subsubsection{Acknowledgments}
This work was supported by a JAE-Predoc fellowship from CSIC, and the research grants: 2009-SGR-1434 from the Generalitat de Catalunya, CSD2007-0022 from MICINN, and Next-CBR TIN2009-13692-C03-01 from MICINN. We thank the reviewers for their constructive input.

\bibliographystyle{splncs03}
\bibliography{ICCBR2011MediationBaydin}
\end{document}